\documentclass{article}

\usepackage{PRIMEarxiv}

\usepackage[utf8]{inputenc} 
\usepackage[T1]{fontenc}    
\usepackage{hyperref}       
\usepackage{url}            
\usepackage{booktabs}       
\usepackage{amsfonts}       
\usepackage{nicefrac}       
\usepackage{microtype}      
\usepackage{lipsum}
\usepackage{fancyhdr}       
\usepackage{graphicx}       
\usepackage{amsmath}
\usepackage{multirow}
\graphicspath{{media/}}     

\title{Sentiment and Emotion Classification of Epidemic Related Bilingual data from Social Media 
}

\author{
  Muhammad Zain Ali\\
  Al-Khawarizmi Institute of Computer Sciences\\
  University of Engineering \& Technology \\
  Lahore, Pakistan\\
  \texttt{muhammad.zain@kics.edu.pk} \\
  \And
  Kashif Javed\\
  Department of Electrical Engineering \\
  University of Engineering \& Technology \\
  Lahore, Pakistan\\
  \texttt{kashif.javed@uet.edu.pk}\\
  \AND
    Ehsan ul Haq\\
  Al-Khawarizmi Institute of Computer Sciences\\
  University of Engineering \& Technology \\
  Lahore, Pakistan\\
  \texttt{ehsan.ulhaq@kics.edu.pk} \\
    \And
  Anoshka Tariq\\
  Department of Electrical Engineering \\
  University of Engineering \& Technology \\
  Lahore, Pakistan\\
  \texttt{anoshkatariq193@gmail.com}
  
}

\begin{document}
\maketitle

\begin{abstract}
In recent years, sentiment analysis and emotion classification are two of the most abundantly used techniques in the field of Natural Language Processing (NLP). Although sentiment analysis and emotion classification are used commonly in applications such as analyzing customer reviews, the popularity of candidates contesting in elections, and comments about various sporting events; however, in this study, we have examined their application for epidemic outbreak detection. Early outbreak detection is the key to deal with epidemics effectively, however, the traditional ways of outbreak detection are time-consuming which inhibits prompt response from the respective departments. Social media platforms such as Twitter, Facebook, Instagram, etc. allow the users to express their thoughts related to different aspects of life, and therefore, serve as a substantial source of information in such situations. The proposed study exploits the bilingual (Urdu and English) data from Twitter and NEWS websites related to the dengue epidemic in Pakistan, and sentiment analysis and emotion classification are performed to acquire deep insights from the data set for gaining a fair idea related to an epidemic outbreak. Machine learning and deep learning algorithms have been used to train and implement the models for the execution of both tasks. The comparative performance of each model has been evaluated using accuracy, precision, recall, and f1-measure. Based on the acquired accuracy measure, the performance of deep learning models was found out to be better than the performance of machine learning models. Besides, the comparative performance of one-vs-rest and multiclass classifiers is also analyzed in this paper and expressed in terms of the weighted average of precision and weighted average of recall. Our results indicated that the multiclass classifiers had a higher value of weighted recall, whereas, the one-vs-rest classifiers had a higher value of weighted precision. 
\end{abstract}


\section{Introduction}\label{sec:intro}
Sentiment polarity classification and emotion classification are two hot topics of research regarding Natural Language Processing (NLP) in general and text classification in particular, which are being used abundantly to get some really smart insights about every aspect of life from customers’ feedback about a certain product to predicting electoral results. Different lexical or machine learning techniques are normally used to determine sentiment polarity, which can give information about a user’s statement being positive, negative or neutral in the semantic sense. Similarly, emotion classification is done to determine the underlying emotion category in a given statement, text or document. It usually starts with a selection of an emotion model which provides the available emotion categories in which the data understudy would fall. Despite the growing popularity of sentiment and emotion classification for analyzing customer reviews and various other applications, their use for the outbreak detection or disease surveillance is relatively unexplored. 

Recently it has been seen that a growing number of people are suffering from different epidemics every year. The inability to deal with the situation efficiently, renders devastating outcomes. One of the main reasons of the increased number of causalities is ineffective surveillance. The key elements to deal with the epidemics is timely action and effective policy making by the government and health departments. However, the traditional means of outbreak detection are inefficient and time consuming which inhibits prompt response management. Poor surveillance means in Pakistan have resulted in many causalities due to different epidemics. Dengue, in particular, has been affecting Pakistan since 2010, resulting in almost 20,000 positive cases and claiming more than 300 lives all over the country \cite{who19s}. Previously, a concerted effort was made for the development of such a dedicated system capable of acquiring and integrating data from multiple sources to deal with different epidemics throughout the year \cite{ali2016id}. However, the proposed system inherently lagged behind the trigger point of the outbreak as it relied on the registered data of the patients in hospitals. Therefore, a much quicker source of data is required for early outbreak detection.

It is an immovable fact that an average person- if feeling unwell- doesn't consult a health specialist unless he feels severity in his condition, however, there is a strong possibility that he will search about his condition on the internet or post that on social media platforms to find the possible cure. The rapid growth in the use of smart phones and increased availability of mobile broadband internet over the years have made such practices very common. People are expressing their thoughts, NEWS and daily life activities on social media platforms in the language of their choice. As a result, social media platforms such as Twitter, Facebook, Instagram and Snapchat etc. have gained tremendous amount of popularity in recent years, and therefore can serve as a viable source of information for various survey related applications. Hence, the proper analysis of such large chunk of data may provide us with the required disease surveillance and outbreak detection. 

In this study, we acquired bilingual (Urdu and English) disease related data from Twitter and NEWS websites. The reason of choosing Twitter as the data source for the sake of this study is that its limited words feature provides comparative convenience in text preprocessing and features extraction. Using the collected data, we performed sentiment analysis and emotion classification by employing machine learning algorithms such as Linear Space Vector Machines (LSVM) \cite{cortes1995support} and Multinomial Naïve Bayes’ \cite{kibriya2004multinomial}, and deep learning algorithms such as 1-d Convolutional Neural Networks (CNN) \cite{lecun1995convolutional}, Long Short-Term Memory (LSTM) networks \cite{hochreiter1997long}, Bidirectional LSTM \cite{zhou2016text} and Attention Mechanism with LSTM \cite{bahdanau2014neural}. Based on the acquired values of accuracy and F1-score, the performance of deep learning algorithms was found out to be way better than the machine learning algorithms.  Additionally, we have also explored the comparative performance of two different categorical classification approaches for both tasks i.e. one-vs-rest and multiclass and expressed in terms of the weighted average of precision and weighted average of recall. The results indicated that the multiclass classifiers had a higher value of weighted recall, whereas, the one-vs-rest classifiers had a higher value of weighted precision.

Although, there has been a considerable work in the domain of sentiment analysis and emotion classification as it constitutes a promising area of research but very less work has been done in processing the data shared on social media in Urdu language. It has also been observed that a growing number of people are communicating and expressing their thoughts in Urdu alphabets which has produced a big room for the researchers to pursue the domain of sentiment analysis and emotion classification for this language. To the best of our knowledge, this is the first work which exploits both machine learning and deep learning algorithms for sentiment analysis and emotion classification of epidemic-related bilingual (Urdu and English) data and also compares the performance of one-vs-rest and multiclass categorical classifiers.
    
The rest of the paper is organized as follows. In Section \ref{sec:rw}, previous works related to the proposed work have been discussed. In Section \ref{sec:mip}, methodology of data set collection, data labeling and model building is presented. In Section \ref{sec:pe}, performance evaluation parameters are defined. In Section \ref{sec:rnc}, comparative performance of resulting models are evaluated using accuracy, precision, recall and f1-measure. Conclusions and future dimensions of this research have been presented in Section \ref{sec:cfs}.
\section{Related work} \label{sec:rw}
Sentiment analysis and emotion classification are two of the most researched areas of Natural Language Processing (NLP). In text classification, the previous works follow either lexical approach or machine learning approach for determination of sentiment or emotion category of an unknown sample. Majority of the previous work in sentiment analysis and emotion classification has been done in the domains of product reviews, sports, movie reviews and blogs. 

Jain et al. presented similar works in their papers on disease surveillance using data from social media \cite{jain2018effective}\cite{jain2016novel}. However, the classification tasks were carried out using machine learning algorithms such as SVM and Naïve Bayes’ which didn’t quite produce optimal results. Therefore, the main focus in this paper will be to explore the application of deep learning algorithms on the collected data set to improve the result. Do et al. \cite{do2019} presented a comprehensive overview of deep learning techniques used for aspect based sentiment analysis. Based on the gathered results, it was concluded that pre-trained and fine-tuned word embedding vectors, incorporation of linguistic factors such as part-of-speech and grammatical rules, and exploration of concept-based knowledge exhibit considerable improvement in the results. Similarly, Liu et al. \cite{liu2017multi} performed multi-class sentiment classification using feature selection and machine learning algorithms on various datasets. Their main contribution is the conduction of over three thousand experiments and documentation of the comparative results. Becker et al. \cite{becker2017multilingual}, on the other hand, investigated emotion classification in English and Portuguese language based on Ekman’s model of basic emotions \cite{ekman1992argument}. They postulated that the use of multilingual data improves the performance of the emotion classification models to a considerable amount. In \cite{singh2016role}, an interesting approach to deal with the slang words in tweets for sentiment analysis has been discussed. Authors calculated joint probability of slang words with actual words to update scores of a given sample. Their experiment suggested a notable improvement in the overall results.

Similarly, a substantial amount of Urdu language based work has been done in the domain of sentiment classification using both lexical and machine learning approaches. A simple approach for determining sentiment polarity has been presented in \cite{amjad2017exploring}. Authors developed Urdu sentiment lexicon and assigned sentiment polarity to the sample based on the accumulated score of the individual words. Asif et al. \cite{asif2020sentiment} performed sentiment analysis to classify multilingual textual data into extremism categories. Their contribution was the development of multilingual sentiment lexicon with the intensity weights. However, they used Linear SVM and Multinomial Naïve Bayes’ for model training, which are known to have inconsistent performance on different datasets for tasks like sentiment analysis. In \cite{mahmood2020deep}, authors used N-gram model, baseline model and recurrent convolutional neural network model for sentiment classification of Roman Urdu data. They found the performance of RCNN to be better than the other two. Contrary to the results proving the dominance of machine learning approaches for sentiment classification, Mukhtar et al. \cite{mukhtar2018lexicon} hypothesized that the lexical approach has superior performance over supervised machine learning approach in their work based on Urdu language data belonging to various genres. However, in a broader perspective, lexical methods cannot possibly capture the underlying semantic meaning of text due to inconsideration of the linguistic factors, which results in subpar performance for sentiment analysis. Hasan et al. \cite{hasan2018machine} used machine learning algorithms for performing sentiment analysis on bilingual twitter data. They translated the Urdu tweets into English and then used lexical libraries for sentiment polarity labeling of entire data set, before training their models. Nevertheless, such approach may only be as good as lexical approach since the data is labeled using lexical libraries.
\section{Methodology}\label{sec:mip}
\subsection{Data Collection \& Processing} \label{sec:motiv}
\begin{figure}[ht]
	\centering
		\includegraphics[scale=0.65]{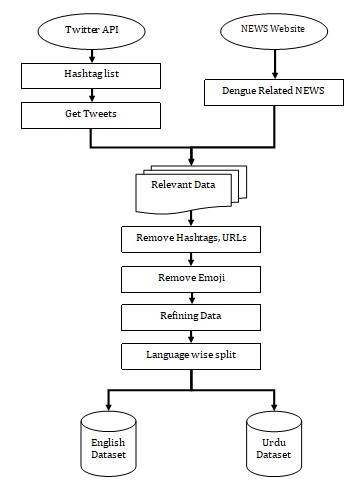}
	\caption{Data Collection Process}
	\label{fig:data_coll}
\end{figure}

As discussed earlier, we made use of Twitter API and NEWS websites for gathering a comprehensive set of bilingual text samples. Figure \ref{fig:data_coll} shows the sequence of steps followed for the acquisition of required data set.
After setting up the twitter developer account, we passed on the hashtag list consisting of dengue related key words (e.g. `dengue’, `fever’, `break bone’, `aedes aegypti’ etc.) in both English and Urdu to the Twitter API to get the required tweets. Similarly, we acquired dengue related data from various NEWS websites as well. Since dengue fever was at its peak in Pakistan during August and September 2019, we collected the raw data for these two months only. During the collection process, we only kept the data consisting of information related to symptoms, preventions or NEWS about the outbreak. After getting a substantial amount of raw data, we applied initial text preprocessing to remove hashtags, URLs and emojis from the entire data set. During the manual refining process, we discarded irrelevant samples and corrected spelling or grammatical mistakes. Finally, we split the dataset into English and Urdu subsets. After following the mentioned series of steps and final filtration, we were able to acquire a total of 5044 samples with equal proportion of both languages. We used 70-30 train test split ratio for experimentation in this project.
\subsection{Assigning Sentiment Labels}
Labeling the sentiment polarity of a given text sample is quite an ambiguous task in NLP. A sentence may appear to be of different polarities to different people. We conducted a series of experiments for labeling the sentiment polarities of the collected samples to find the optimal method. Firstly, we did content analysis to assign polarity labels to the samples based on selected positive and negative words in them. Secondly, we used TextBlob \cite{loria2018textblob} and SentiWordNet \cite{baccianella2010sentiwordnet} lexical libraries to assign sentiment scores to each sample. For Urdu data subset, we used Translation API of Google Cloud to convert the samples into English before passing them to the lexical libraries. However, these two approaches didn’t give the correct sentiment polarity distribution and therefore, falsely reflected the public’s sentiments.
 
\begin{figure}[h]
	\centering
		\includegraphics[scale=0.6]{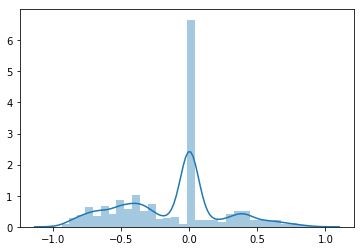}
	\caption{Compound Score Distribution of Data set}
	\label{fig:comp_score}
\end{figure}

\begin{figure}[h]
	\centering
		\includegraphics[scale=0.6]{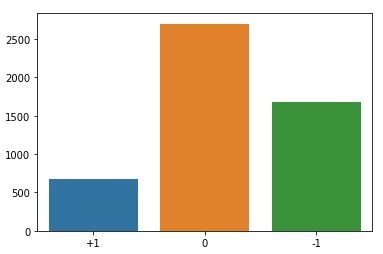}
	\caption{Sentiment Polarity Count}
	\label{fig:sent_score}
\end{figure}

To overcome the mentioned problem, we used \textit{SentimentIntensityAnalyzer} function of NLTK Vader lexicon library \cite{loper2002nltk}, which returns the compound score calculated by summation of individual words’ lexicon ratings and then normalized between +1 and -1. The resultant distribution plot of compound scores of entire data set is shown in Figure \ref{fig:comp_score} which resembles a sinc function.
The threshold values for positive and negative sentiment polarities were selected to be +0.3 and -0.3 respectively. Based on these values, the resultant count plot of assigned polarities is shown in Figure \ref{fig:sent_score}, which is quite an accurate representation since most of the samples doesn't bear any sentiment while negative sentiments prevail the positive ones in such a situation. 
\subsection{Assigning Emotion Labels}\label{sec:ael}
\begin{figure}[h]
	\centering
		\includegraphics[scale=0.7]{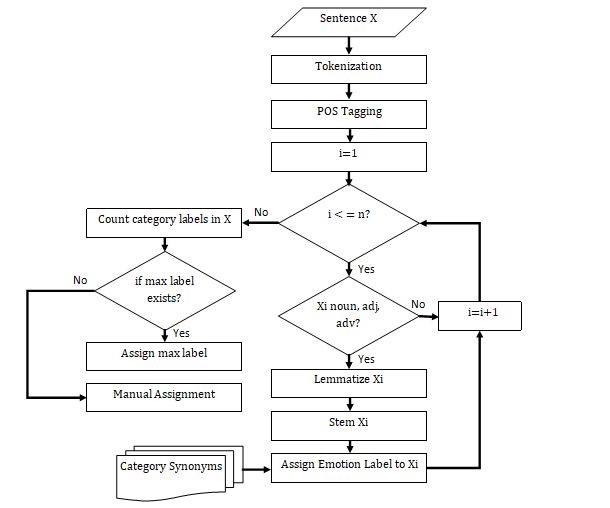}
	\caption{Baseline System for Assigning Emotion Labels}
	\label{fig:emo_labelling}
\end{figure}

\begin{figure}[h]
	\centering
		\includegraphics[scale=0.6]{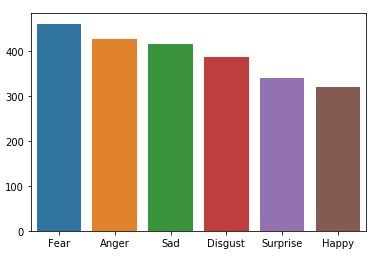}
	\caption{Emotion Labels Distribution}
	\label{fig:emo_count}
\end{figure}

The task of assigning emotion labels to the data is linked with sentiment polarities as only positive and negative sentiments can be assigned emotion labels, whereas neutral sentiments don’t contain any emotion. We selected Ekman’s model \cite{ekman1992argument} for assigning emotion labels which consists of six categories i.e. `happiness’, `sadness’, `surprise’, `anger’, `fear’ and `disgust’. The baseline system designed to assign emotion label to a given positive or negative sentiment bearing sentence is shown in Figure \ref{fig:emo_labelling}. 
First of all, a given sentence X is broken down into a list of its own words, where n represents the total number of words in that sentence. Each individual word is then tagged with the part of speech using \textit{pos\textunderscore tag} function of NLTK Vader lexicon library \cite{loper2002nltk}. The system iterates over each word in the sentence and only considers noun, adjective or adverb for further process. The lemma and stem of the noun, adjective or adverb tagged word is assigned category based on its presence in either of the six predefined list objects for each emotion category containing all the synonyms from the English Thesaurus. After repeating the process for all the words in the sentence, the emotion categories assigned to the individual words are summed up. The system then assigns the dominant category as an overall emotion label to the sentence. For the cases where dominant category was missing, we manually accessed the content and assigned final emotion label to the sentence. We repeated the same practice for Urdu data samples after translating them to English using Translate API of Google Cloud. The resulting bar plot of the assigned emotion labels to the collected samples using the above mentioned approach is shown in Figure \ref{fig:emo_count} which is presumably an apt depiction of public’s emotions since most of the people possess the feeling of fear in such situations. 
\begin{figure}[h]
	\centering
		\includegraphics[scale=0.6]{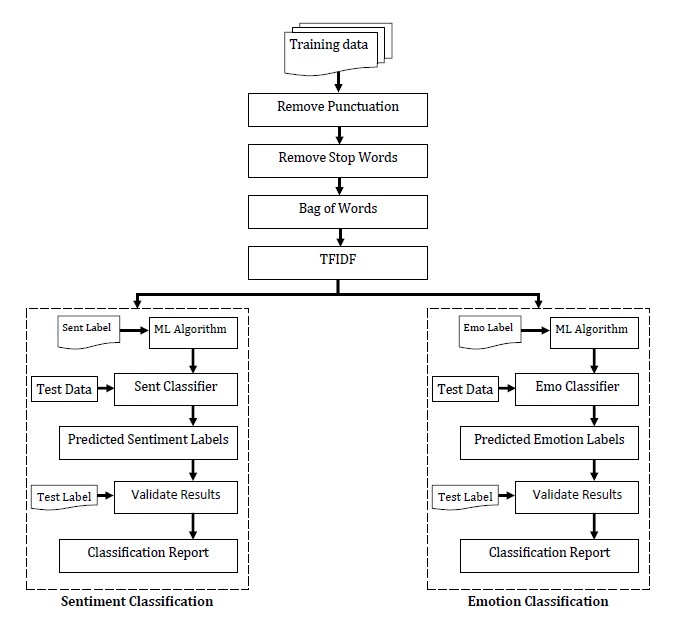}
	\caption{Model Building using Machine Learning Algorithms}
	\label{fig:mlmb}
\end{figure}
\subsection{Model Building using Machine Learning Algorithms}\label{sec:mbmla}
In this study we have used two of the most commonly used machine learning algorithms for model building i.e. Multinomial Naïve Bayes’ and Linear Space Vector Machines. MNB makes use of probability theory and Bayes’ theorem with assumption of naïve independence among the features to learn from the data set and predict the class of unknown inputs. SVM, on the other hand, works on the basis of maximal margin classifier. It is basically an optimization problem where we have to maximize the margin between the decision boundary (or hyperplane) and the nearest lying training samples in the feature space. Figure \ref{fig:mlmb} shows the block diagram of machine learning model building process that we employed in this study. 
We built two dedicated pipelines for handling English and Urdu samples separately. Since text data is unstructured by nature, therefore, we had to perform a series of steps to remove unnecessary information and extract features from the training data set. First of all, punctuation symbols were removed. After that, we performed content analysis to remove stop words from the data set. Stop words is basically a term in NLP which refers to the most commonly used words in a given language. The reason for removing the stop words is that their contribution to put a given sample to a certain class is negligible. For English samples, we used stop words from NLTK corpus \cite{loper2002nltk} and for Urdu samples we used the stop words list provided in \cite{jabbar2016analysis}. After that, we used Bag-of-words technique which extracts unique words from the data set and represent each sample as a row of frequency table of those unique words. Thus, each row is a sample and each column represents the number of times a given word occurred in that sample, thereby, creating a sparse matrix representation for each sample. After that, we applied TFIDF(term frequency-inverse document frequency), which is basically a technique used to scale the features according to their existence in a single document as well as in the entire data set. Finally, we fed the extracted features to the machine learning algorithms for training the classifiers in order to execute the sentiment and emotion classification tasks. The newly trained classifiers were then fed the extracted features of the test data and we validated the predicted classes against the actual labels. The classification performance report of the employed algorithms for both the tasks have been documented in the Section \ref{sec:rnc} of this paper. 
\subsection{Model Building using Deep Learning Algorithms}\label{sec:mbdla}
\begin{figure}[ht]
	\centering
		\includegraphics[scale=0.65]{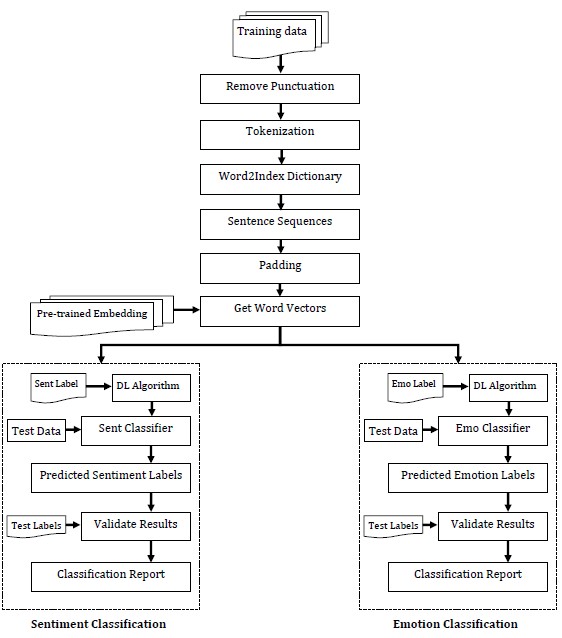}
	\caption{Model Building using Deep Learning Algorithms}
	\label{fig:dlmb}
\end{figure}

In this study we have used 1-d CNN, LSTM, Bidirectional LSTM and attention mechanism with LSTM for the execution of sentiment classification and emotion classification tasks. Convolutional neural network is known to have an outstanding performance for image data classification due to its resemblance with biological visual cortex in living organisms. Recent studies have suggested its remarkable performance for text classification tasks as well \cite{collobert2011natural}\cite{kim2014convolutional}. Long Short-Term memory models, on the other hand, are useful for the cases where the order of data or historical information is important. Similarly, bidirectional LSTM incorporates a forward and a backward layer to learn from the tokens on both sides. Additionally, attention mechanism is a process employed to address the issue of irrelevant information encoding in conventional encoder-decoder framework such as recurrent neural network. To make use of these algorithms for the execution of sentiment and emotion classification tasks, the sequence of steps which we followed are shown in Figure \ref{fig:dlmb}. 
Unlike machine learning models, we don’t have to do a lot of feature engineering for deep learning model building. First of all, we removed the punctuation symbols from the training data set. Secondly, we tokenized the data and created word to index dictionary for assigning numbers to the unique words in the data set. Using the acquired dictionary object, we represented each training sample as a list of indices of its constituent words. After that, we padded required number of zeros with each acquired list to make them equal to an even length of 100. In the next step, we created an embedding matrix of the data set using pre-trained word vectors. The size of the embedding matrix was $n \times d$, where $n$ represents the number of unique words in the data set and $d$ represents the embedding dimension which was selected as 100 in our case. We used word2vec \cite{mikolov2013efficient} and fastText \cite{grave2018learning} pre-trained vectors for English and Urdu data set respectively. Each word in the word to index dictionary object was looked up in either of the two pre-trained vectors list and the acquired word vector was updated in the embedding matrix row specified by the index of that word in the dictionary object. Additionally, we had to represent the sentiment and emotion labels in one hot encoded vector form. Thus, using sentence sequence representation, hot encoded labels and acquired embedding matrix, we were able to train deep learning models for sentiment and emotion classification tasks. Finally, the trained classifiers were provided sentence sequences of the test data and the predicted classes were validated against the actual hot encoded labels of the test data. The classification performance reports have been documented and discussed in the Section \ref{sec:rnc}.
\section{Performance Evaluation}\label{sec:pe}
The performance of the proposed models is measured based on Accuracy (Acc), Precision (P), Recall (R) and F-score (F1). For binary class labels, these scores are given by: 
\begin{align}
&P = \frac{TP}{TP+FP}\\
&R = \frac{TP}{TP+FN}\\
&F1=\frac{2PR}{P+R}
\end{align}
\begin{equation}
Acc = \frac{TP+TN}{TP+TN+FP+FN}
\end{equation} 
where TP (true postive), TN (true negative), FP (False positive) and FN (False negative) represent the number of correct or incorrect predictions by the system based on actual labels. Precision measures the percentage of correctly predicted labels in a given class with respect to the total predictions made by the system in that class, whereas, Recall measures the percentage of correctly predicted labels by the system in a given class with respect to the actual number of labels in that class. F-score is a balance between the mentioned two measures and is computed by taking the harmonic mean of the precision and recall. On the other hand, the Accuracy measure represents the overall performance of the entire system.
\section{Results}\label{sec:rnc}
In this section we have presented the performance evaluation results of both machine learning and deep learning methodologies employed in this study. In order to avoid any bias in the performance results, we took five 70-30 random splits of the entire data set for training and testing of each model. We have published the average of the 5 resulting classification reports for each model in the subsequent sections.
\subsection{Sentiment Classification}
\subsubsection{Performance of Machine Learning Algorithms}
By employing the methodology presented in Section \ref{sec:mbmla}, we trained our machine learning models for sentiment classification of our data. Multinomial Naïve Bayes’ classifier resulted in an average accuracy of 71\% for English samples and 68.8\% for Urdu samples. On the other hand, using SVM, we got a very high accuracy of 80\% for English samples and 74.6\% for Urdu samples. The detailed performance reports for both algorithms are shown in Table \ref{t1} and Table \ref{t2} for English and Urdu data sets respectively.
\begin{table}[h]
\centering
\begin{tabular}{lccc|ccc}
\hline
\multicolumn{7}{c}{Average Performance for English Data set}\\\hline
\multicolumn{4}{c|}{Multinomial Naïve Bayes'} & \multicolumn{3}{c}{SVM} \\\hline
Categories & Precision & Recall & F1-score & Precision & Recall & F1-score\\\hline
Negative & 0.764 & 0.702 & 0.732 & 0.860 & 0.792 &  0.826\\
Neutral & 0.680 & 0.880 & 0.766 & 0.784 & 0.882 & 0.828\\
Positive & 0.984 & 0.076 & 0.142 & 0.732 & 0.508 & 0.596\\\hline
\end{tabular}
\caption{Sentiment Classification using SVM $\&$ MNB for English Samples}
\label{t1}
\end{table}
\begin{table}[ht]
\centering
\begin{tabular}{lccc|ccc}
\hline
\multicolumn{7}{c}{Average Performance for Urdu Data set}\\\hline
\multicolumn{4}{c|}{Multinomial Naïve Bayes'} & \multicolumn{3}{c}{SVM} \\\hline
Categories & Precision & Recall & F1-score & Precision & Recall & F1-score\\\hline
Negative & 0.772 & 0.640 & 0.696 & 0.822 & 0.704 &  0.756\\
Neutral & 0.654 & 0.890 & 0.756 & 0.716 & 0.880 & 0.786\\
Positive & 0.922 & 0.054 & 0.106 & 0.724 & 0.356 & 0.476\\\hline
\end{tabular}
\caption{Sentiment Classification using SVM $\&$ MNB for Urdu Samples}
\label{t2}
\end{table}
It can be seen that SVM based sentiment classifier performed better with higher average accuracy and f1-score values for both English and Urdu data sets. However, the recall value for positive class was observed to be low especially for MNB. It is because of the reason that the positive samples were less compared to the other classes which resulted in underfitting for the positive class in general. Secondly, sentiment classification is a complicated task and probability-based models such as MNB may not entirely capture the underlying semantics in order to predict the right label.
\subsubsection{Performance of Deep Learning Algorithms}
As discussed earlier, we used 1-d CNN, LSTM, Bidirectional LSTM and attention mechanism for training our models by employing methodology presented in Section \ref{sec:mbdla}. Using 1-d CNN, we got an average accuracy of 71\% for English data set and 65\% for Urdu data set. Similarly, LSTM model gave an average accuracy of 76\% for English data set and an average accuracy of 70.8\% for Urdu data set. The detailed performance reports are tabulated in Table \ref{t3} and Table \ref{t4} for English and Urdu data respectively.
\begin{table}[h]
\centering
\begin{tabular}{lccc|ccc}
\hline
\multicolumn{7}{c}{Average Performance for English Data set}\\\hline
\multicolumn{4}{c|}{1-d CNN} & \multicolumn{3}{c}{LSTM} \\\hline
Categories & Precision & Recall & F1-score & Precision & Recall & F1-score\\\hline
Negative & 0.736 & 0.712 & 0.716 & 0.846 & 0.784 &  0.814\\
Neutral & 0.726 & 0.774 & 0.744 & 0.784 & 0.794 & 0.786\\
Positive & 0.586 & 0.436 & 0.492 & 0.576 & 0.590 & 0.556\\\hline
\end{tabular}
\caption{Sentiment Classification using CNN $\&$ LSTM for English Samples}
\label{t3}
\end{table}
\begin{table}[ht]
\centering
\begin{tabular}{lccc|ccc}
\hline
\multicolumn{7}{c}{Average Performance for Urdu Data set}\\\hline
\multicolumn{4}{c|}{1-d CNN} & \multicolumn{3}{c}{LSTM} \\\hline
Categories & Precision & Recall & F1-score & Precision & Recall & F1-score\\\hline
Negative & 0.704 & 0.706 & 0.702 & 0.740 & 0.718 &  0.726\\
Neutral & 0.686 & 0.692 & 0.688 & 0.716 & 0.794 & 0.752\\
Positive & 0.390 & 0.364 & 0.368 & 0.592 & 0.384 & 0.456\\\hline
\end{tabular}
\caption{Sentiment Classification using CNN $\&$ LSTM for Urdu Samples}
\label{t4}
\end{table}
\begin{table}[!h]
\centering
\begin{tabular}{lccc|ccc}
\hline
\multicolumn{7}{c}{Average Performance for English Data set}\\\hline
\multicolumn{4}{c|}{Bidirectional LSTM} & \multicolumn{3}{c}{Attention Mechanism} \\\hline
Categories & Precision & Recall & F1-score & Precision & Recall & F1-score\\\hline
Negative & 0.834 & 0.814 & 0.823 & 0.848 & 0.892 &  0.870\\
Neutral & 0.808 & 0.844 & 0.825 & 0.894 & 0.838 & 0.866\\
Positive & 0.636 & 0.516 & 0.567 & 0.660 & 0.764 & 0.706\\\hline
\end{tabular}
\caption{Sentiment Classification using enhanced LSTM models for English Samples}
\label{t5}
\end{table} 
\begin{table}[!h]
\centering
\begin{tabular}{lccc|ccc}
\hline
\multicolumn{7}{c}{Average Performance for Urdu Data set}\\\hline
\multicolumn{4}{c|}{Bidirectional LSTM} & \multicolumn{3}{c}{Attention Mechanism} \\\hline
Categories & Precision & Recall & F1-score & Precision & Recall & F1-score\\\hline
Negative & 0.765 & 0.772 & 0.770 & 0.796 & 0.810 &  0.800\\
Neutral & 0.749 & 0.816 & 0.779 & 0.822 & 0.814 & 0.812\\
Positive & 0.636 & 0.400 & 0.469 & 0.598 & 0.624 & 0.610\\\hline
\end{tabular}
\caption{Sentiment Classification using enhanced LSTM models for Urdu Samples}
\label{t6}
\end{table}

Keeping in view the achieved values of average accuracy and f1-score, LSTM model turned out to be better than 1-d CNN for sentiment classification. We got an eminent increase in accuracy measure using bidirectional LSTM and attention mechanism. Using bidirectional LSTM, the accuracy values increased up to 79.6\% and 74.4\% for English and Urdu data sets respectively. Similarly, using attention mechanism with bidirectional LSTM, the accuracy values reached 84.4\% and 78.4\% for English and Urdu data sets respectively. The detailed performance reports consisting of precision, recall and f1-score for both English and Urdu data sets are shown in Table \ref{t5} and Table \ref{t6} respectively.
Based on the acquired values of average accuracy and f1-score, it is evident that the bidirectional LSTM with attention mechanism outperformed the rest of the models discussed above for sentiment classification task.
\subsection{Emotion Classification}
\subsubsection{Performance of Machine Learning Algorithms}
By using the identical set of features and algorithms as used for sentiment classification task and the emotion labels annotated by employing the methodology discussed in Section \ref{sec:ael}, we trained machine learning models for emotion classification task. Using MNB, we got an average accuracy values of 61.5\% and 57.6\% for English and Urdu data samples respectively. Similarly, the average accuracy values for SVM based emotion classifier were found out to be 72.6\% and 73.6\% for English and Urdu data samples respectively. The detailed performance report of these models for both English and Urdu samples are shown in Table \ref{t7} and Table \ref{t8}. 
\begin{table}[h]
\centering
\begin{tabular}{lccc|ccc}
\hline
\multicolumn{7}{c}{Average Performance for English Data set}\\\hline
\multicolumn{4}{c|}{Multinomial Naïve Bayes'} & \multicolumn{3}{c}{SVM}\\\hline
Categories & Precision & Recall & F1-score & Precision & Recall & F1-score\\\hline
Anger&0.515&0.670&0.585&0.680&0.743&0.707\\
Disgust&0.530&0.495&0.510&0.693&0.720&0.703\\
Fear&0.660&0.540&0.590&0.703&0.660&0.683\\
Happy&0.780&0.675&0.715&0.803&0.737&0.763\\
Sad&0.600&0.665&0.620&0.727&0.760&0.737\\
Surprise&0.735&0.720&0.720&0.813&0.763&0.783\\\hline
\end{tabular}
\caption{Emotion Classification using SVM \& MNB for English Samples}
\label{t7}
\end{table}
\begin{table}[ht]
\centering
\begin{tabular}{lccc|ccc}
\hline
\multicolumn{7}{c}{Average Performance for Urdu Data set}\\\hline
\multicolumn{4}{c|}{Multinomial Naïve Bayes'} & \multicolumn{3}{c}{SVM}\\\hline
Categories & Precision & Recall & F1-score & Precision & Recall & F1-score\\\hline
Anger&0.550&0.570&0.557&0.733&0.723&0.727\\
Disgust&0.623&0.380&0.470&0.767&0.670&0.710\\
Fear&0.470&0.630&0.537&0.627&0.720&0.673\\
Happy&0.733&0.527&0.613&0.843&0.710&0.773\\
Sad&0.527&0.577&0.547&0.747&0.767&0.750\\
Surprise&0.700&0.807&0.747&0.767&0.847&0.807\\\hline
\end{tabular}
\caption{Emotion Classification using SVM \& MNB for Urdu Samples}
\label{t8}
\end{table}

The overall performance of these models was found out to be quite similar to that of the sentiment classification models. Based on the average values of accuracy and f1-score, it can be seen that the SVM trained model outperformed the MNB trained model for emotion classification task as well.

\subsubsection{Performance of Deep Learning Algorithms}
Following the same set of steps as followed to train sentiment classification deep learning models, we trained deep learning models for emotion classification of positive and negative sentiment bearing data samples as well. Using 1d-CNN, we got average accuracy values of 74\% and 73.6\% for English and Urdu data samples respectively. Similarly, using LSTM, we got average accuracy values of 75.6\% and 74.7\% for English and Urdu data samples respectively. The detailed performance reports consisting of precision, recall and f1-score values are tabulated in Table \ref{t9} and Table \ref{t10}. 
\begin{table}[h]
\centering
\begin{tabular}{lccc|ccc}
\hline
\multicolumn{7}{c}{Average Performance for English Data set}\\\hline
\multicolumn{4}{c|}{1-d CNN} & \multicolumn{3}{c}{LSTM}\\\hline
Categories & Precision & Recall & F1-score & Precision & Recall & F1-score\\\hline
Anger&0.835&0.705&0.760&0.787&0.747&0.767\\
Disgust&0.645&0.845&0.730&0.683&0.757&0.717\\
Fear&0.720&0.710&0.710&0.737&0.737&0.733\\
Happy&0.820&0.755&0.785&0.830&0.807&0.813\\
Sad&0.735&0.650&0.690&0.710&0.683&0.700\\
Surprise&0.755&0.795&0.775&0.797&0.803&0.800\\\hline
\end{tabular}
\caption{Emotion Classification using CNN \& LSTM for English Samples}
\label{t9}
\end{table}
\begin{table}[h]
\centering
\begin{tabular}{lccc|ccc}
\hline
\multicolumn{7}{c}{Average Performance for Urdu Data set}\\\hline
\multicolumn{4}{c|}{1-d CNN} & \multicolumn{3}{c}{LSTM}\\\hline
Categories & Precision & Recall & F1-score & Precision & Recall & F1-score\\\hline
Anger&0.657&0.733&0.693&0.647&0.697&0.667\\
Disgust&0.827&0.716&0.768&0.637&0.737&0.687\\
Fear&0.672&0.672&0.672&0.797&0.657&0.717\\
Happy&0.795&0.671&0.728&0.877&0.827&0.847\\
Sad&0.663&0.695&0.679&0.717&0.747&0.730\\
Surprise&0.826&0.897&0.860&0.897&0.867&0.880\\\hline
\end{tabular}
\caption{Emotion Classification using CNN \& LSTM for Urdu Samples}
\label{t10}
\end{table}
Moreover, using bidirectional LSTM, we got average accuracy values of 77.6\% and 76.4\% for English and Urdu data sets respectively. Attention mechanism, on the other hand, gave average accuracy values of 76\% and 75.3\% for English and Urdu data sets respectively. The detailed performance reports of Bidirectional LSTM and attention mechanism for English and Urdu data sets are shown in Table \ref{t11} and Table \ref{t12}.
\begin{table}[ht]
\centering
\begin{tabular}{lccc|ccc}
\hline
\multicolumn{7}{c}{Average Performance for English Data set}\\\hline
\multicolumn{4}{c|}{Bidirectional LSTM} & \multicolumn{3}{c}{Attention Mechanism}\\\hline
Categories & Precision & Recall & F1-score & Precision & Recall & F1-score\\\hline
Anger&0.847&0.767&0.800&0.727&0.790&0.750\\
Disgust&0.817&0.727&0.763&0.757&0.727&0.740\\
Fear&0.727&0.740&0.730&0.660&0.643&0.653\\
Happy&0.823&0.817&0.820&0.817&0.830&0.820\\
Sad&0.723&0.790&0.750&0.793&0.780&0.790\\
Surprise&0.767&0.830&0.800&0.827&0.837&0.837\\\hline
\end{tabular}
\caption{Emotion Classification using enhanced LSTM models for English Samples}
\label{t11}
\end{table}
\begin{table}[h]
\centering
\begin{tabular}{lccc|ccc}
\hline
\multicolumn{7}{c}{Average Performance for Urdu Data set}\\\hline
\multicolumn{4}{c|}{Bidirectional LSTM} & \multicolumn{3}{c}{Attention Mechanism}\\\hline
Categories & Precision & Recall & F1-score & Precision & Recall & F1-score\\\hline
Anger&0.665&0.715&0.687&0.655&0.699&0.676\\
Disgust&0.655&0.755&0.697&0.639&0.741&0.686\\
Fear&0.815&0.675&0.737&0.805&0.662&0.726\\
Happy&0.895&0.845&0.867&0.878&0.829&0.853\\
Sad&0.735&0.765&0.747&0.725&0.748&0.736\\
Surprise&0.914&0.885&0.897&0.902&0.871&0.886\\\hline
\end{tabular}
\caption{Emotion Classification using enhanced LSTM models for Urdu Samples}
\label{t12}
\end{table}
Based on the acquired values of average accuracy and f1-score, it can be seen that bidirectional LSTM outperformed the rest of the models for emotion classification task for both English and Urdu data sets.
\clearpage
\subsection{Comparison of One-vs-rest \&  Multiclass Classifiers}
\begin{figure}[ht]
	\centering
		\includegraphics[scale=0.5]{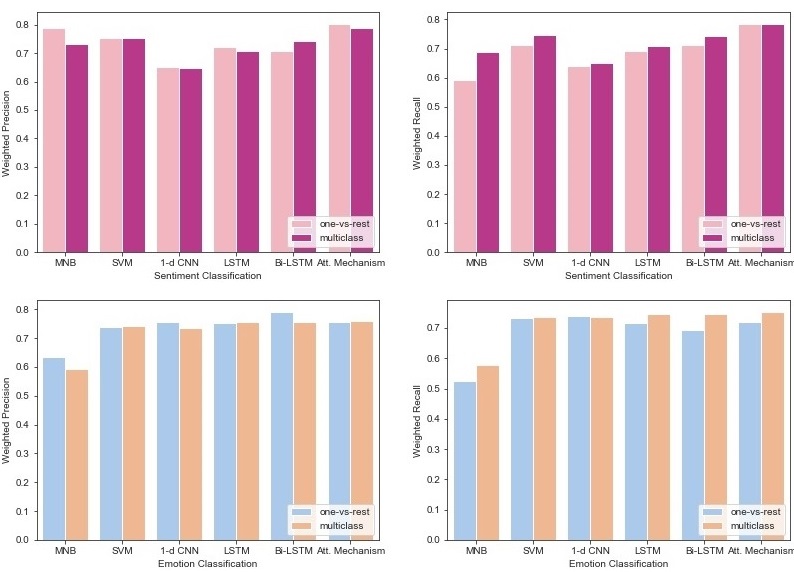}
	\caption{Comparison between One-vs-rest \& Multiclass classifiers for English}
	\label{fig:onrvmed}
\end{figure}

\begin{figure}[hb]
	\centering
		\includegraphics[scale=0.5]{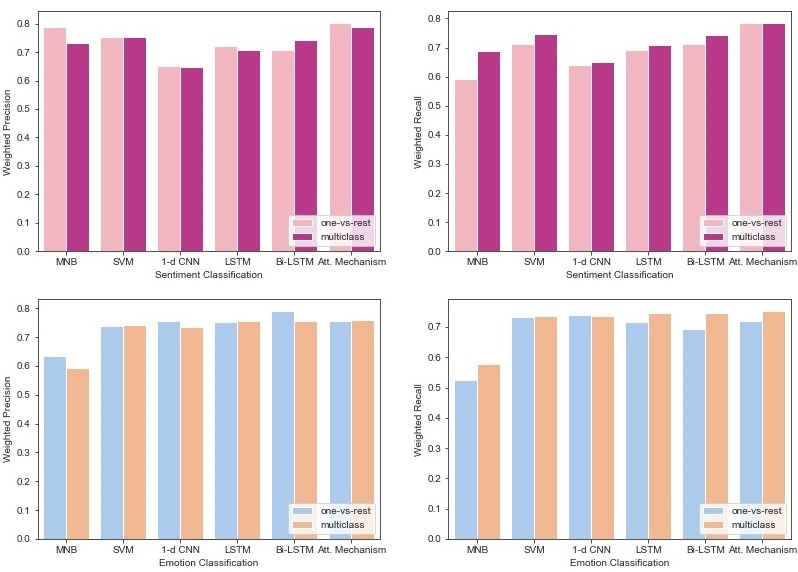}
	\caption{Comparison between One-vs-rest \& Multiclass classifiers for Urdu}
	\label{fig:onrvmud}
\end{figure}

In addition to the comparison of performance of machine learning and deep learning algorithms, we also analyzed the performance of two different approaches for categorical classification of both tasks i.e. one-vs-rest and multiclass classification. One-vs-rest refers to the type of classification where we train a single binary classifier per class, with the samples of only that class positive while all the other samples as negative. On the other hand, multiclass classifiers are simply the extension of binary classifiers for three or more class labels. The results of these classification techniques- in terms of weighted precision and weighted recall- are shown in Figure \ref{fig:onrvmed} and Figure \ref{fig:onrvmud} for English and Urdu data samples respectively.  
The overall trend of the bar plots show that one-vs-rest models had higher values of weighted precision while multiclass models had higher values of weighted recall. It implies that one-vs-rest models yielded higher percentage of correct predictions with respect to the total predictions made by them, whereas, multiclass models had relatively higher percentage of correct predictions with respect to the actual labels of the data set.
\subsection{Discussion of Results}
The summary of task-wise performance of machine learning and deep learning algorithms in terms of accuracy is displayed in Table \ref{t13}. Sentiment classification is a complex task and thus requires deeper insights into the extracted features to predict the correct results. A simple machine learning algorithm like Multinomial Naïve Bayes’ which rely only on the conditional probability, can't possibly capture the underlying semantic information for sentiment and emotion classification tasks. That is why, the performance of MNB was observed to be particularly poor for these tasks, especially the recall score for positive sentiments. Although, the performance of SVM with grid search was found out to be consistent but the performance of deep learning algorithms especially the Bidirectional LSTM model with attention mechanism outperformed the rest of the models in sentiment classification domain. 
The overall performance of deep learning algorithms was found out to be better than machine learning algorithms for emotion classification task as well. However, simple bidirectional LSTM performed better than bidirectional LSTM with attention mechanism. The fact that the performance of sentiment classification models was better than emotion classification models is due to the reason that the number of class categories increased while the overall data got less because of discarding `neutral’ sentiment bearing data samples during emotion annotation and model training. As a result, less data density caused relative underfitting and hence, lower accuracy of deep learning emotion classifying models compared to sentiment classifying deep learning models. Finally, the performance comparison of one-vs-rest and multiclass classifiers indicated a relatively higher accuracy value of multiclass classifiers as shown in Figure \ref{fig:onrvmed} and Figure \ref{fig:onrvmud}. This is due to the reason that the size of our data is unsuitable for training binary classifiers in one-vs-rest case, as the number of positive instances become significantly less during training. It results in underfitting of the binary models for each class, thereby, causing a notable decrease in the weighted recall value.  
\begin{table}[ht]
\centering
\begin{tabular}{c|l|l|l}
\hline
\multirow{2}{*}{Task} & \multirow{2}{*}{Classifier} & \multicolumn{2}{c}{\% Accuracy} \\\cline{3-4}
 & & English & Urdu \\\hline
\multirow{6}{*}{Sentiment Classification} & MNB & 71.0 & 68.8 \\
 & SVM & 80.0 & 74.6 \\
 & 1-d CNN & 71.0 & 65.0 \\
 & LSTM & 76.6 & 70.8 \\
 & Bi-LSTM & 79.6 & 74.4 \\
 & Att. Mechanism & 84.4 & 78.4 \\\hline
\multirow{6}{*}{Emotion Classification} & MNB & 61.5 & 57.6 \\
 & SVM & 72.6 & 73.6 \\
 & 1-d CNN & 74.0 & 73.6 \\
 & LSTM & 75.6 & 74.7 \\
 & Bi-LSTM & 77.6 & 76.4 \\
 & Att. Mechanism & 76.0 & 75.3\\\hline
\end{tabular}
\caption{Summary of Task-wise Models' Performance}
\label{t13}
\end{table}

\section{Conclusion and future scope}\label{sec:cfs}
In this study, an NLP based approach has been presented to track public sentiments and emotions related to epidemics using English and Urdu data shared on social media. The purpose is to observe the stats and the stream of data related to epidemic to gain a fairly well reasoned idea about the trigger point of an outbreak. After acquiring relevant bilingual data, sentiment labels were assigned by dynamically choosing the cut-off scores acquired from NLTK lexical library. Similarly, emotion annotations were assigned by doing content analysis of the collected samples. Many preprocessing techniques and state-of-the-art machine learning and deep learning algorithms were investigated to acquire the most optimal results. Overall performance of deep learning algorithms was found out to be better than the machine learning algorithms for both the tasks. Besides, the comparative performance of one-vs-rest and multiclass classification models was also analyzed which rendered multiclass classification approach to be more accurate than the other one. 

With a growing trend of people using the Urdu alphabets to express their opinions on social media, it has become extremely important to produce some high quality and rich resources for Urdu language like generation of extensive Urdu word vectors and sentiment lexicons based on entire Urdu dictionary. This research can be further pursued by acquiring more data from multiple resources such as blogs, forums and other social media platforms. Another interesting thing can be the incorporation of emoticons and stickers for sentiment analysis and emotion classification tasks.
\bibliographystyle{unsrt}  
\bibliography{references}

\end{document}